%% file: main.tex
\pdfoutput=1 % This is recommended for using pdfLaTeX
\documentclass[11pt]{article}

\usepackage[table]{xcolor} % Automatically loads 'colortbl'
\usepackage{times}
\usepackage[T1]{fontenc}
\usepackage[utf8]{inputenc}
\usepackage{graphicx}
\usepackage{subfig}
\usepackage{amsmath,amsfonts}
\usepackage{algpseudocode,algorithm}
\usepackage{array}
\usepackage{threeparttable}
\usepackage{adjustbox}
\usepackage{booktabs}
\usepackage{placeins}
\usepackage{xcolor}
\definecolor{Gray}{gray}{0.9}
\definecolor{brightturquoise}{rgb}{0.85, 1, 1}

\usepackage{hyperref}

\usepackage{acl}

\usepackage{microtype}

\title{Towards Hierarchical 
Spoken Language Disfluency Modeling}

\author{Jiachen Lian \\
  UC Berkeley\\
  \texttt{jiachenlian@berkeley.edu} \\\And
  Gopala Anumanchipalli \\
  UC Berkeley\\
  \texttt{gopala@berkeley.edu} \\}

\begin{document}
\maketitle
\input{abstract}

\input{introduction}

\input{method}

\input{experiments}

\input{conclusion}

\bibliography{reference}

\clearpage
\onecolumn
\appendix

\input{appen}

\end{document}

%% file: abstract.tex
\begin{abstract}
Speech disfluency modeling is the bottleneck for both speech therapy and language learning. However, there is no effective AI solution to systematically tackle this problem. We solidify the concept of \textit{disfluent speech} and \textit{disfluent speech modeling}. We then present \textit{Hierarchical Unconstrained Disfluency Modeling} (H-UDM) approach, the hierarchical extension of~\citet{lian2023unconstrained-udm} that addresses both disfluency transcription and detection to eliminate the need for extensive manual annotation. Our experimental findings serve as clear evidence of the effectiveness and reliability of the methods we have introduced, encompassing both transcription and detection tasks.
\end{abstract}

%% file: introduction.tex
\section{Introduction}

Spoken language disfluency modeling is the core technology in speech therapy and language learning. According to~\citet{nidcd}, an estimated \textit{17.9 million} adults and \textit{1.4 percent} of children in the U.S. suffer from chronic communication and speech disorders. Currently, hospitals have to invest substantial resources in hiring speech and language pathologists (SLPs) to manually analyze and provide feedback. More importantly, the cost is not affordable for low-income families. Kids' speech disorders also have a significant connection to the language learning market. According to a report by~\citet{vcl}, the English language learning market will reach an estimated value of \textit{54.8 billion} by 2025. Unfortunately, there is not an AI tool that can effectively automate this problem.

\begin{figure*}[ht]
    \centering
    \includegraphics[width=15cm, height=8cm]{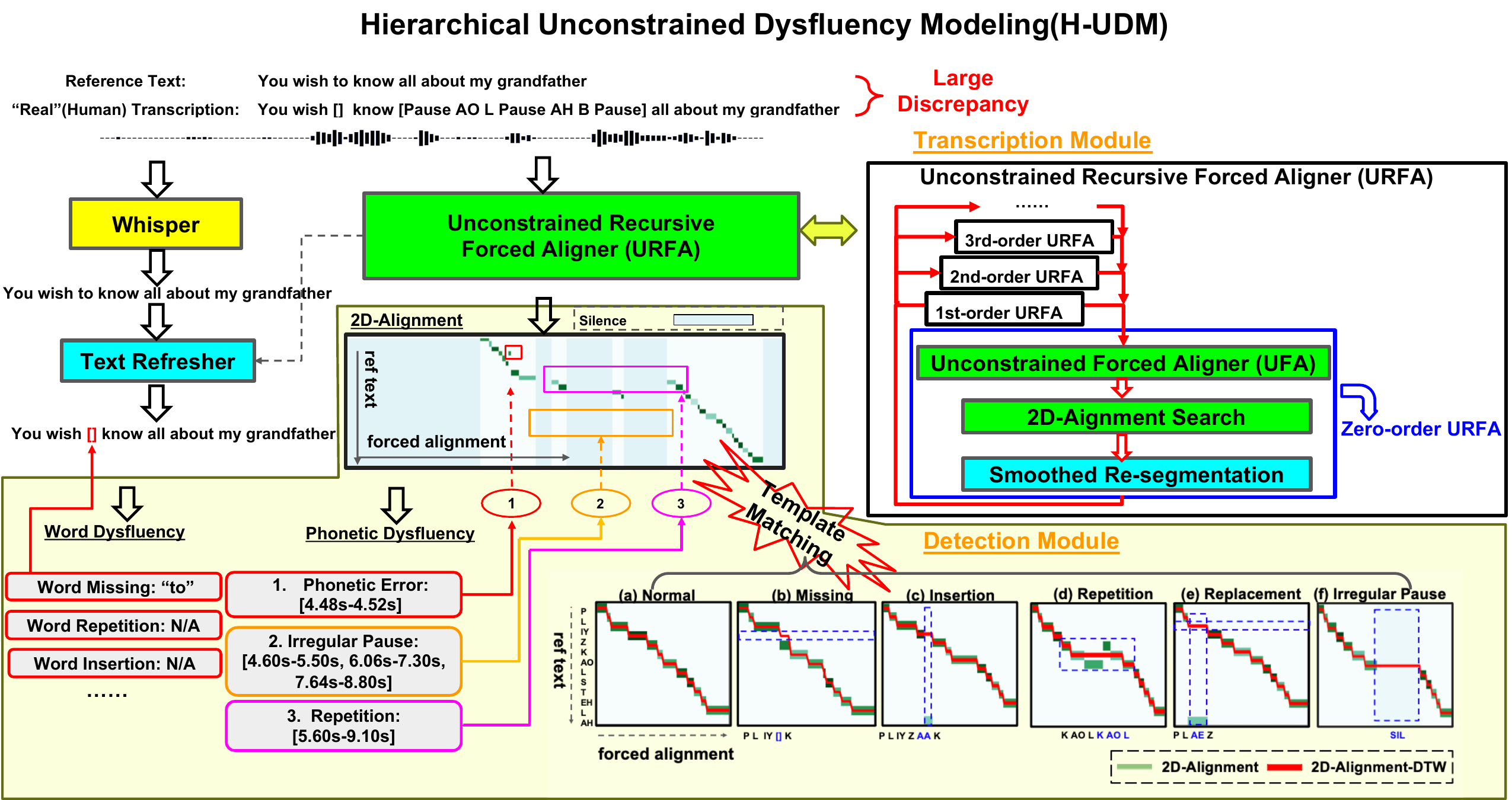}
    \caption[]{Hierarchical Unconstrained disfluency Modeling(H-UDM) consists of \textit{Transcription} module and \textit{Detection} module. Both word-level and phoneme-level disfluencies are detected and localized. Here is an example of aphasia speech. The reference text is "You wish to know all about my grandfather," while the real/human transcription differs significantly from the reference. Whisper~\cite{radford2022whisper} recognizes it as perfect speech, while H-UDM is able to capture most of the disfluency patterns. An audio sample of this can be found here\protect\footnotemark.}
    \label{UDM}
\end{figure*}

In current research community, there is not a unified definition for \textit{disfluent speech}, as mentioned in~\citet{lian2023unconstrained-udm}. As such, we solidify the definition of  \textit{disfluent speech} as any form of speech characterized by abnormal patterns such as repetition, replacement, and irregular pauses, as summarized in~\citet{lian2023unconstrained-udm}. 
Within the domain of \textit{disfluent speech modeling}, research efforts are conducted both on the speech side and the language side. Whenever disfluent speech transcription is given (such as \textit{human transcription} in Figure \ref{UDM}), the problem can be tackled by LLMs~\cite{openai2023gpt4}. However, such transcription is not available and current best ASR systems such as~\citet{radford2022whisper} tend to recognize them as perfect speech. Thus, we argue that the bottleneck lies in the \textit{speech side} rather than in language. 

Unfortunately, there is also no established definition for the problem of speech disfluency modeling. We \textit{formally define} that \textit{speech disfluency modeling} is to detect all types of disfluencies at both the word and phoneme levels while also providing a time-stamp for each type of disfluency. In other words, disfluency modeling should be \textbf{hierarchical} and \textbf{time-accurate}. Previous research has mainly focused on a small aspect of this problem.

Researchers started by focusing on spotting stuttering using end-to-end methods. They manually tagged each utterance and developed the classification model at the utterance level~\cite{kourkounakis2021fluentnet, alharbi2017segment-detection2, alharbi2020segment-detection3, segment-detection4}. Later on, things got detailed with frame-level stutter detection~\cite{harvill2022frame-level-stutter, shonibare2022frame-detection2}. However, end-to-end methods have their limitations. First, stuttering is just one aspect of disfluency. Current end-to-end models struggle to handle other forms of disfluency effectively. Second, manually labeling data for these methods is a lot of work and not practical for larger-scale projects. Lastly, disfluency modeling depends on the specific text being spoken, a factor that has been overlooked in previous research, as pointed out in~\citet{lian2023unconstrained-udm}. 

It is typically intuitive to consider speech transcription that offers disfluency-specific representations. For a long time, the mainstream of researchers in speech transcription has been focused on word-level automatic speech recognition (ASR), which has been further scaled. However, The most advanced word transcription models currently available~\cite{radford2022whisper, zhang2023google-usm, pratap2023scaling-speech, aghajanyan2023scaling-speech, lian2023av-data2vec} can only transcribe certain obvious word-level disfluency patterns, such as word repetition or replacement. However, the majority of disfluencies occur at the phoneme-level or subword-level, making them challenging for any ASR system to explicitly detect. ~\citet{kouzelis2023weakly} introduced a neural forced aligner that incorporates time accuracy and sensitivity to silence. This aligner employs a weighted finite-state transducer (WFST) to capture disfluency patterns like repetition. However, it fails on openset disfluency modeling~\cite{lian2023unconstrained-udm}.

The Unconstrained Disfluency Model (UDM) introduced in~\cite{lian2023unconstrained-udm} was devised to address the aforementioned challenges comprehensively. UDM seamlessly integrates both transcription and detection modules within a unified framework. Within the UDM framework, non-monotonic alignments are acquired through dynamic alignment search, forming the foundation for subsequent template matching algorithms aimed at detecting various disfluency types. Specifically, distinct templates are tailored for each disfluency category, encompassing replacements, insertions, deletions, blocks, and repetitions. Additionally, VCTK++ dataset was introduced to further enhance model performance. In the present study, we extend the capabilities of UDM by incorporating a monotonicity constraint. While non-monotonic alignment is essential for effective disfluency modeling, our experiments demonstrate that the integration of a simple Connectionist Temporal Classification (CTC) module alongside a phoneme classifier can enhance non-monotonicity. Furthermore, we introduce the \textit{Unconstrained Recursive Forced Aligner} (URFA), which employs an iterative process to generate both phoneme alignments (1D) and 2D alignments with weak text supervision. This recursive modeling significantly enhances detection robustness. Our proposed method, termed \textit{Hierarchical Unconstrained Disfluency Modeling} (H-UDM), attains state-of-the-art performance in real aphasia speech disfluency detection.

\begin{figure*}[ht]
    \centering
    \includegraphics[width=16cm, height=6.5cm]{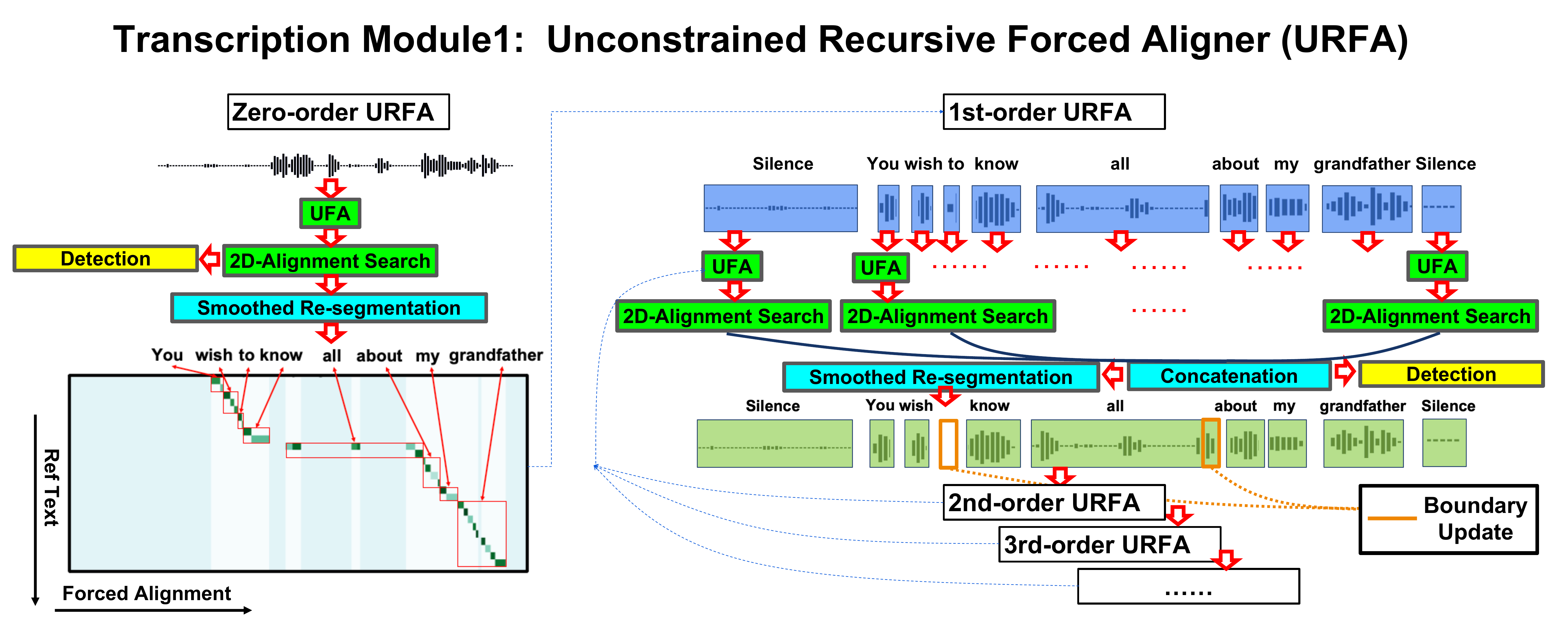}
    \caption[]{Unconstrained Recursive Forced Aligner consists of three basic modules: \textit{UFA}, \textit{2D alignment Search}, \textit{Smoothed Re-segmentation}. In the first iteration (Zero-order), the entire utterance is taken and 2D alignment is generated. Starting at 2nd iteration (1st-order), the disfluent speech is segmented at word level and each segment is processed separately and then combined to generate the final 2D alignment for detection. }
    \label{URFA}
\end{figure*}

\footnotetext{Fig.\ref{UDM} Audio samples. (1) Aphasia Speech Sample: \url{https://shorturl.at/eTWY1}. (2) Template speech samples: \url{https://shorturl.at/bszVX}}

%% file: method.tex
\section{Transcription Module}
Our transcription module consists of two core parts: \textit{(1) Unconstrained Recursive Forced Aligner}, which generates phonetic transcriptions (2D-Alignment), and \textit{(2) Text Refresher} which takes both Whisper output and 2D-Alignment to generate word transcription, as shown in Fig. \ref{UDM}.

\subsection{Unconstrained Recursive Forced Aligner} \label{URFA-sec}
The bottleneck for disfluent speech alignment is that the \textit{real text transcription} is unknown, which is significantly different from the \textit{reference text}, as shown in Fig. \ref{UDM}. However, disfluency detection relies on the \textit{reference text}. Traditional speech-text aligners~\cite{mcauliffe2017montrealmfa, kim2021conditional-vits, li2022neufa} assume that the \textit{reference text} is the same as the \textit{real text transcription}, and thus they only work for normal fluent speech. Let's look at a simple example. If the \textit{reference text} is "K AE Y (Cat)" and the actual speech (\textit{real text transcription}) is "K AE K AE T (Ca-Cat)," then the alignment from traditional aligners will all be "K AE T" as monotonicity is enforced, which is not accurate. For disfluent speech detection, deriving non-monotonic speech-text alignment is required, and this is achieved through the Unconstrained Forced Aligner (UFA)~\cite{lian2023unconstrained-udm}. As disfluency detection depends on the \textit{reference text}, we also introduce 2D-Alignment to align the non-monotonic phoneme alignment with the \textit{reference text}. Additionally, we deploy our alignment methods recursively, re-segmenting the utterance based on the 2D-Alignment to refine 2D-Alignment itself. The entire paradigm is illustrated in Fig. \ref{URFA}. Each sub-module is detailed in the following.

\subsubsection{UFA}

The Unconstrained Forced Aligner (UFA)~\cite{lian2023unconstrained-udm} operates by predicting alignments with the guidance of weak text supervision. Initially, the speech segment undergoes encoding by the WavLM \cite{chen2022wavlm} encoder, which generates latent representations. Subsequently, a conformer module \cite{gulati2020conformer} is employed to predict both alignment and boundary information. The alignment and boundary targets used in UFA are derived from the Montreal Forced Aligner (MFA) \cite{mcauliffe2017montrealmfa}. During the inference stage, there is no requirement for text input, rendering the alignment process truly "unconstrained." To perform phoneme classification, UFA simply applies two linear layers. For the phoneme classifier, UFA optimizes the softmax cross-entropy objective, while logistic regression is utilized for boundary prediction. Notably, we found through experimentation that introducing an additional Connectionist Temporal Classification (CTC) constraint~\cite{graves2006connectionist-ctcs} (monotonicity) can enhance the robustness of our non-monotonic alignment. It's important to emphasize that CTC is solely involved in the training stage. For more in-depth model details, please refer to~\citet{lian2023unconstrained-udm}.

\paragraph*{Dynamic Alignment Search}  \label{DAS}
We adopt the alignment search methodology proposed by~\citet{lian2023unconstrained-udm}. It is essential to note that, in the context of disfluency modeling, the alignment must be non-monotonic. This stands in stark contrast to traditional forced aligners, which typically enforce monotonic alignment based on supervised signals such as text. However, in our case, text supervision is complicated by the substantial divergence between the real transcription and the reference text. Consequently, the reference text becomes an unreliable source for alignment. The process of decoding the alignment sequence from the emission matrix can be accomplished through various methods. In our approach, we follow the methodology outlined in~\citet{lian2023unconstrained-udm} and apply the boundary-aware Viterbi algorithm for decoding. It is worth noting that the modified Viterbi algorithm introduces a computational complexity of O($tN^2$), where $N$ represents the vocabulary size and $t$ denotes the number of time steps. Given that, in practice, $t$ is typically much larger than $N$, this computational complexity remains within acceptable bounds. The inclusion of boundary information proves invaluable in handling the ambiguity introduced, particularly by silence.
In addition, we trained a phoneme autoregressive language model using the VCTK corpus~\cite{yamagishi2019cstr-vctk}. Nevertheless, we did not observe a significant improvement in performance. Therefore, we opted to adhere to the approach outlined in~\citet{lian2023unconstrained-udm} and continued to utilize the bi-gram model.
For a more comprehensive understanding of the search algorithm, please refer to \citet{lian2023unconstrained-udm}.

\subsubsection{2D-Alignment Modeling}
The concept of \textit{2D-Alignment} was initially introduced in~\citet{lian2023unconstrained-udm}. The underlying idea revolves around a fundamental question: how accurately does the forced alignment correspond to the reference text? The 2D-Alignment was devised as a metric to assess this alignment. Specifically, the 2D-Alignment represents the temporal alignment between the actual spoken text by the speaker (ground truth text) and the disfluent alignment generated by the dynamic alignment search module. In the work presented in~\citet{lian2023unconstrained-udm}, this 2D alignment was computed by performing element-wise multiplication between the reference phoneme embeddings and the forced alignment phoneme embeddings. It is important to note that this 2D-Alignment is inherently non-monotonic. However, this approach has significant limitations. Through real speech testing, we observed that in the presence of noise, the noise can become erroneously aligned with parts of the reference text, which is not desirable. Additionally, using phonemes as the primary units for disfluency modeling may not be optimal. For example, there may be minimal phonetic distinctions between certain phonemes, such as 'AH' and 'AO,' in terms of verbal pronunciation. Nonetheless, in both non-monotonic alignment and 2D-Alignment, they are treated as distinct phonemes and are considered uncorrelated. Despite these limitations, we still retained the ground truth 2D-Alignment for template matching algorithms. This ground truth 2D-Alignment, known as \textit{2D-Alignment-DTW}, is always monotonic in nature. In the following section, we will delve into our strategies for addressing the aforementioned challenges.

 \begin{figure}[h]
    \centering
    \includegraphics[height=4.5cm, width=7cm]{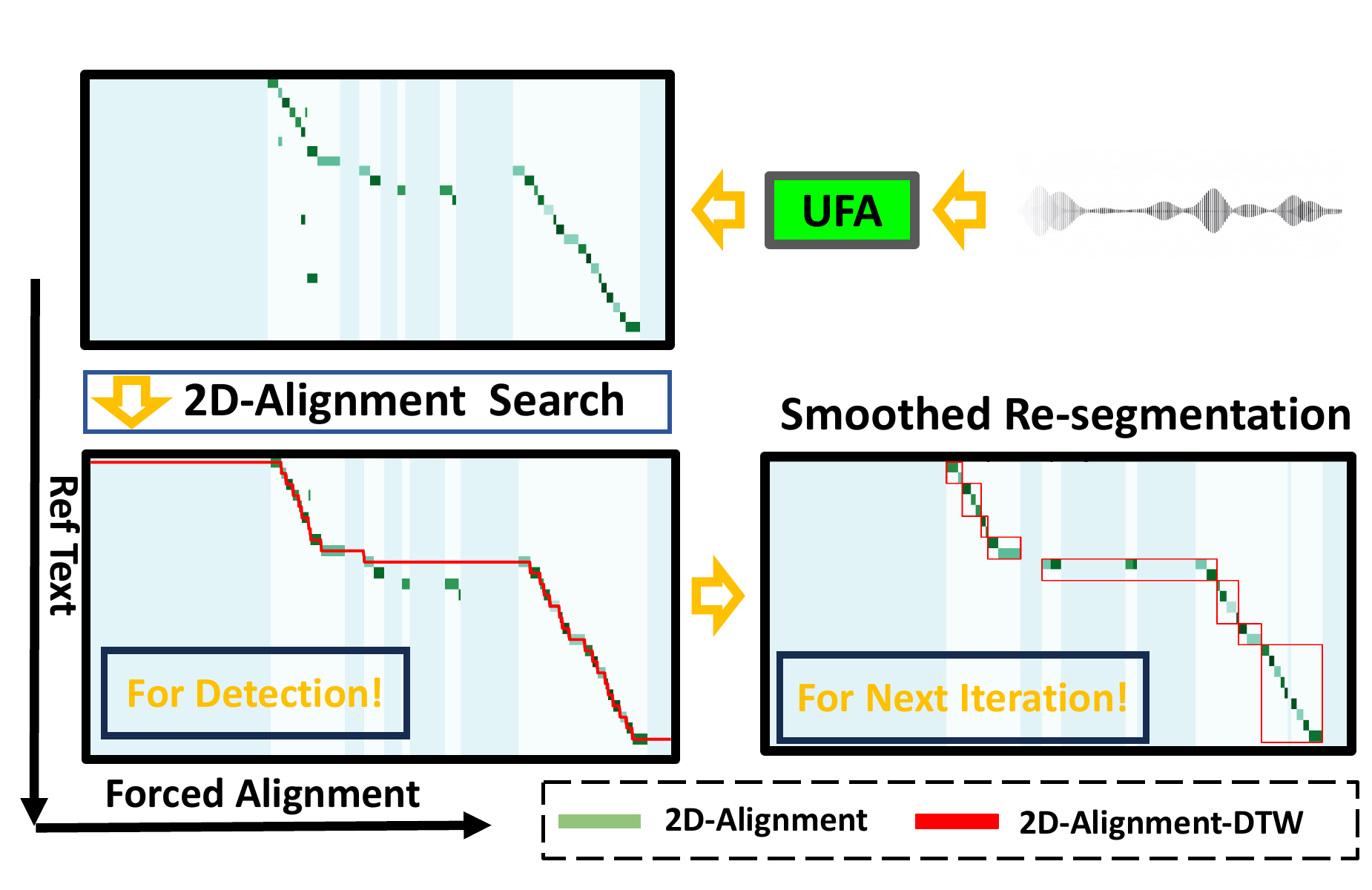}
    \caption{2D-Alignment Modeling}
    \label{2dalign-fig}
\end{figure}
\vspace{-5mm}
\paragraph*{Smoothed Re-segmentation and Recursive Alignment}
The generation of non-monotonic alignment inherently introduces variances that can lead to misdetection. To address this issue, we propose segmenting the disfluent speech by word boundaries and generating alignment for each segment, potentially mitigating the problem. For instance, consider the case illustrated in Fig. \ref{UDM} and Fig. \ref{URFA}, where the sequence [AO L Pause AH B] actually corresponds to the word "all." Another source of variance arises when individuals utter sequences like "AH, AO, AY," which may indicate the repetition of the phoneme "AH." However, our 2D alignment treats them as distinct phonemes, failing to detect the repetition, which poses a significant challenge. To tackle this issue, we introduce a phoneme smoothing technique. Specifically, at each time step, we calculate the cosine similarity of phoneme embeddings for both 2D-Alignment and 2D-Alignment-DTW. If the similarity falls within a predefined threshold, we merge the 2D-Alignment into 2D-Alignment-DTW, as demonstrated in the final figure of Fig. \ref{2dalign-fig}. This process yields a monotonic 2D alignment, allowing us to identify word boundaries by simply locating each word along the "ref text" axis. These segmented results serve as input for 1st-order Unconstrained Forced Aligner (URFA), as depicted in Fig. \ref{URFA}. In 1st-order URFA, we compute a 2D-Alignment for each segment and subsequently concatenate them. This iterative approach can be extended to 2nd-order URFA, 3rd-order URFA, and beyond. It is important to note that the smoothed monotonic 2D-Alignment is exclusively used for segmentation purposes, while the original non-monotonic 2D-Alignment remains in use for detection. This recursive aligner yields improved word boundary detection, as exemplified in Fig. \ref{URFA}, where the boundaries obtained in 1st-order alignment outperform those of zero-order alignment in capturing disfluencies.

\subsection{ASR Scalability}
 \begin{figure}[h]
    \centering
    \includegraphics[height=5cm, width=8cm]{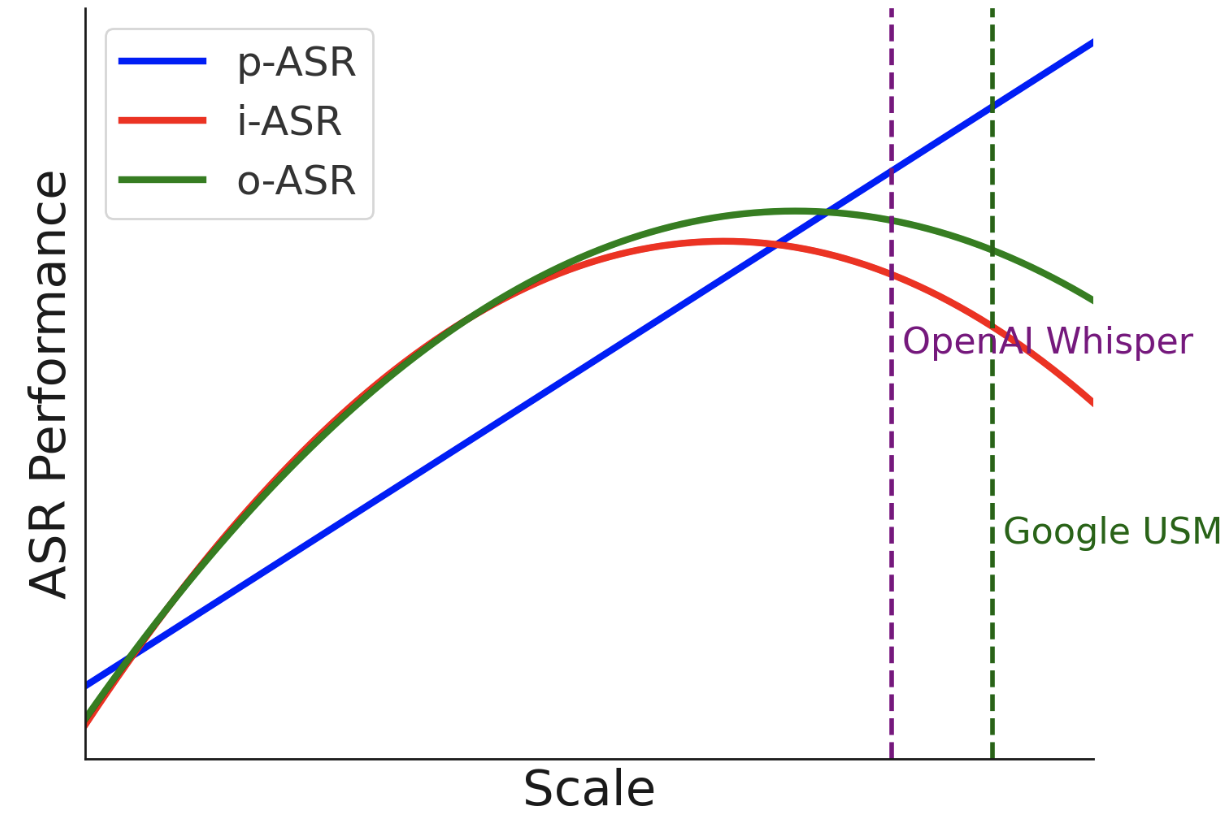}
    \caption{Scaling law for ASR under various conditions. (i) Perfect ASR (p-ASR); (ii) Imperfect ASR(i-ASR); (iii) Overall ASR(o-ASR)}
    \label{i-asr}
\end{figure}

Recent advances in spoken language processing~\cite{pratap2023scaling-speech, aghajanyan2023scaling-speech, zhang2023google-usm, lian2023av-data2vec} indicate the effectiveness of scaling laws concerning data and model scale. The limit of scaling has not been reached yet. However, the scaling law for ASR is most effective for normal or perfect speech (p-ASR in Fig.\ref{i-asr}). In real-life settings, things are very different for imperfect speech, such as disfluent speech. Due to the power of language modeling in ASR systems, most imperfect speech is treated as perfect speech, leading to a significant performance drop for imperfect ASR (i-ASR in Fig.\ref{i-asr}). The overall ASR (o-ASR in Fig.~\ref{i-asr}), which includes both parts, should also follow the same trend. \citet{lian2023unconstrained-udm} introduced the \textit{text refresher} to introduce imperfections for disfluent speech in an attempt to avoid the aforementioned problems. The solutions are intuitive. Of all imperfections (disfluencies) at the word level, insertions and deletions are the hardest to detect. However, this can be easily observed on the 2D-Alignment introduced in the previous section. In the 2D-Alignment, we also have 2D-Alignment-DTW as a reference. If the 2D-Alignment does not align with any reference words, then it is likely an insertion, and if the word from the ASR system is redundant in comparison to the 2D-Alignment phoneme sequence, it is likely a deletion. It is important to note that URFA also generates word transcriptions. However, based on our findings, it exhibits inferior performance in word-level disfluency detection compared to the "text refresher." Therefore, we have chosen to employ URFA exclusively for phonetic-level disfluency detection

\subsection{Transcription Module Evaluation}

\subsubsection{Duration-Aware Phonetic Transcription}
We follow~\cite{lian2023unconstrained-udm} for phonetic transcription evaluation. Here, we provide more insights for each evaluation metric. First, the transcribed phonemes must be intelligible at the segment level, which is evaluated by the phoneme error rate (\textit{PER}). Second, the transcribed phonemes must be intelligible at the frame-level, which is evaluated by frame-level \textbf{Micro F1 Score} and \textbf{Macro F1 Score}~\cite{f1-sore}. Third, the transcribed phonemes must be intelligible at both the segment and frame levels, which is evaluated by the combination of the above metrics. This is also known as dPER~\cite{lian2023unconstrained-udm}. In more detail, dPER is the duration-aware extension of PER. For each operation to be counted, we consider the duration for it.

\subsubsection{Duration-Aware Imperfect Word Transcription} \label{wer}
Disfluent speech is imperfect speech. Traditional ASR systems are typically evaluated by how well the hypothesis matches the ground truth text. In disfluent settings, ASR systems are evaluated based on how well the hypothesis matches the imperfect targets. We start by following~\cite{lian2023unconstrained-udm} to adopt the imperfect word error rate (i-WER) where the disfluent (imperfect) targets are labeled by humans. In our proposed method, we also employ segment-level imperfect ASR evaluation, similar to dPER vs PER, where duration is also considered. In detail, we calculate the Intersection over Union (IoU) between our predicted time boundaries from URFA and the ground truth boundaries from human annotations. If the IoU is greater than 0.5, the disfluency is identified as detected. We also report the F1 score for this matching evaluation, referred to as the \textbf{Matching Score (MS)}.

\section{Detection Module}
The reason we adopt a separate design for the detection and transcription modules is that an end-to-end modeling approach for the detection system is not reliable. The transcription module provides us with disfluency-aware representations to optimize the detection module. Here, we can still design learning-based methods~\cite{harvill2022frame-level-stutter, shonibare2022frame-detection2, alharbi2017segment-detection2, alharbi2020segment-detection3} to predict the detection results; however, we don't have human labels for disfluencies, which might be considered for future work. Instead, we have developed a smart label-free system that simply employs the template matching algorithm for each type of disfluency. Template matching is efficient and reliable, eliminating the need for human annotation. We have designed disfluency templates for both word and phoneme levels. These disfluencies include \textit{Phonetic Errors (Missing, Deletion, Replacement), Repetition, and Irregular Pause}. Our methods also cover word-level disfluencies, including \textit{Missing, Insertion, Replacement, and Repetition}. The following section details them.

\subsection{Phonetic-Level disfluency Detection}
We follow the approach outlined in~\cite{lian2023unconstrained-udm} for designing disfluency templates. Instead of directly handling the alignment from dynamic alignment search, we also consider alignment data from the URFA module. We repeat the processes described in~\cite{lian2023unconstrained-udm}. In Figure \ref{UDM}-Template, when examining alignments in normal speech, we observe perfect alignment between the two representations. However, closer examination reveals distinctive patterns within these alignments. If we notice a significant drop in alignment-2D-DTW without any overlap in the corresponding row, this signals the presence of a \textbf{missing} phoneme, as depicted in Fig \ref{UDM}-Template-(b). When a row in alignment-2D-DTW intersects with multiple columns in alignment-2D and contains repeated phonemes, it indicates a \textbf{repetition}, as illustrated in Figure \ref{UDM}-template-(d). Conversely, if a row in alignment-2D-DTW aligns with alignment-2D and simultaneously matches the surrounding column in alignment-2D, this signifies an \textbf{insertion}, as exemplified in Figure \ref{UDM}-template(c). When a row in alignment-2D-DTW fails to overlap with any horizontal regions in alignment-2D but does overlap with a single vertical block in alignment-2D, it is categorized as a \textbf{replacement}, as demonstrated in Figure \ref{UDM}-template(e). Lastly, any pauses occurring within a complete sentence are recognized as \textbf{irregular pauses}, as shown in Figure \ref{UDM}-template(f).

\subsection{Word-level disfluency Detection}
We followed the same processes for detecting word-level disfluencies as we did for phoneme-level disfluencies. In line with~\citet{lian2023unconstrained-udm}, neither duration nor silence were taken into consideration. It's important to note that, unlike~\citet{lian2023unconstrained-udm}, we select the best results from either \textit{URFA} or the \textit{text refresher}. We adhere to the evaluation framework proposed by~\citet{lian2023unconstrained-udm} for assessing hierarchical disfluency. To provide a more detailed evaluation, we utilize F1 scores and matching scores that consider temporal labels.

%% file: experiments.tex
\section{Experiments}

\subsection{Datasets and Pre-processing} \label{datasets}
\paragraph*{VCTK~\cite{yamagishi2019cstr-vctk}} We utilize VCTK for training the \textit{UFA} module. We follow the train/test split methodology outlined in~\citet{lian2022robust-d-dsvae, lian2022towards-c-dsvae, lian2022utts, lian2022utts2}.

\paragraph*{VCTK$^{++}$}~\cite{lian2023unconstrained-udm} It is a disfluency-aware simulated speech based on VCTK~\cite{yamagishi2019cstr-vctk}. Three types of disfluencies are introduced: repetitions, prolongations, and blocks. For repetitions and prolongations, phonemes are randomly selected and prolonged or repeated for a random duration. These operations are performed in the temporal domain (waveform). VCTK++ is utilized for training the \textit{UFA}.

\begin{table*}[h]
    \centering
    \setlength{\tabcolsep}{5pt}
    \renewcommand{\arraystretch}{1.2} 
     \resizebox{15cm}{!}{
    \begin{tabular}{l l c |c c c c| c c c c} 
    \toprule
    Method& WavLM Size&Training Data&Micro F1 ($\%$, $\uparrow$) & Macro F1 ($\%$, $\uparrow$) & dPER ($\%$, $\downarrow$) & PER ($\%$, $\downarrow$) &Micro F1 ($\%$, $\uparrow$) & Macro F1 ($\%$, $\uparrow$) & dPER ($\%$, $\downarrow$)  & PER ($\%$, $\downarrow$)\\
     \hline
 \rowcolor{Gray} 
\multicolumn{3}{c}{}& \multicolumn{4}{c}{\textit{Buckeye Test Set}} & \multicolumn{4}{c}{\textit{VCTK++ Test Set}} \\
    WavLM-CTC-VAD&Large&None&50.1&47.3&86.9&12.0&48.8&45.7&88.0&8.2\\
    WavLM-CTC-MFA&Large&None&49.8&28.7&53.9&12.0&47.6&26.0&54.2&8.2\\
    UFA&Base&VCTK&68.9&55.6&53.3&15.0&78.8&59.5&53.4&11.0\\
    UFA&Base&VCTK+Buckeye&65.9&51.6&63.6&16.3&75.2&56.0&60.0&11.8\\
    UFA&Large&VCTK+Buckeye&70.3&55.0&46.2&13.3&80.7&66.4&45.8&11.0\\
    UFA&Large&VCTK&71.3&60.0&46.0&11.9&81.7&72.0&44.0&10.5\\
   \hspace{1em}-- Boundary-aware&Large&VCTK&68.9&52.0&49.9&12.8&78.4&62.9&47.8&10.7\\
   \hspace{1em}+ CTC&Large&VCTK&68.9&52.0&49.9&10.2&78.4&62.9&47.8&7.8\\
UFA&Large&VCTK$^{++}$&73.5&64.0&41.0&11.5&93.6&90.8&38.0&9.2\\
\hspace{1em}-- Boundary-aware&Large&VCTK$^{++}$&71.0&63.7&44.3&12.2&91.1&90.0&42.1&9.6\\
\hspace{1em}+ CTC&Large&VCTK$^{++}$&\textbf{77.2}&\textbf{68.7}&\textbf{40.3}&\textbf{9.5}&\textbf{92.0}&\textbf{90.9}&\textbf{39.8}&\textbf{6.4}\\
    \bottomrule
    \end{tabular}}
    \caption{Phonetic Transcription Evaluation}
    \label{phn-transcription-eval}
\end{table*}

\paragraph*{Buckeye~\cite{pitt2005buckeye}} It includes substantial segments of disfluent speech that have been meticulously annotated with precise time markings. To create our training and testing subsets, we adhere to the methodology outlined in~\cite{lian2023unconstrained-udm}. Buckeye serves as our primary resource for both training the \textit{UFA} module and conducting \textit{Phonetic Transcription Evaluation}. 

\paragraph*{Disorded Speech} \label{Datasets}
We utilize the same corpus as~\citet{lian2023unconstrained-udm}. Collaborating with speech-language pathologists (SLPs), we personally annotate the hierarchical disfluencies. However, since this segment consists of only 20 minutes of aphasia/dyslexia speech, it is exclusively employed for inference purposes. It's important to note that, due to privacy considerations, this particular dataset will not be publicly shared. Please be aware that in the future, we will continue to collect more data from both hospitals and K-5 schools on a larger scale.

% \vspace{-5pt}
\subsection{Phonetic Transcription Experiments}
\citet{lian2023unconstrained-udm} conducted phonetic experiments on several tasks. First, two baselines were attempted. One is named WavLM-CTC-VAD, where VAD introduces silence into the WavLM-CTC alignment. The other is WavLM-CTC-MFA, where phoneme labels from WavLM-CTC\cite{wavlm-ctc} are set as MFA~\cite{mcauliffe2017montrealmfa} targets. Results from \citet{lian2023unconstrained-udm} indicate that UFA outperforms the baselines under various settings (Buckeye test set and VCTK++ test set). In this work, we explore the role of monotonicity that was introduced. Specifically, we applied the CTC constraint to latent embeddings in the UFA module. An additional phoneme recognition module was applied to introduce such monotonicity. The intuition behind introducing this monotonicity is that the learned phonetic alignment still jumps up and down for disfluent speech and is unstable\cite{lian2023unconstrained-udm}. In this module, we only train UFA without any recursive learning, which will be introduced later on. It is worth noting that UFA remains constant throughout the recursive process. Therefore, our evaluation focuses solely on the alignment produced by UFA rather than that of URFA, as the latter is directly proportional to the former. Phonetic transcription results are shown in the Table.\ref{phn-transcription-eval}. 

\subsection{Imperfect Word Transcription Experiments}
\begin{table}[h]
    \centering
    \setlength{\tabcolsep}{10pt}
    \renewcommand{\arraystretch}{1.1} % Adjust the row spacing
    \resizebox{7cm}{!}{
        \small
        \begin{tabular}{l c c c c} 
            \toprule
            &\multicolumn{4}{c}{iWER($\%$, $\downarrow$)} \\
            \hline
            URFA Config & Zero-order  & 1st-order  & 2nd-order & 3rd-order \\
            \hline
            Whisper-Large & 11.3 & - &- &-\\
            +Text Refresher & 9.7 &9.4 &9.2 &9.2\\
            \hspace{1em} +VCTK$^{++}$ & 9.2 &9.0 & 8.7 &8.7 \\
            \hspace{2em} +CTC & \textbf{8.8} & \textbf{8.6} &\textbf{8.4} &\textbf{8.4}\\
            \bottomrule
        \end{tabular}
    }
    \caption{Word Transcription Evaluation}
    \label{word-trans-eval}
\end{table}
We present results from Whisper~\cite{radford2022whisper} and zero-order text refresher from~\cite{lian2023unconstrained-udm}. In these settings, we conduct recursive word transcription modeling in multiple orders. The recursive process involves the following steps: The default UDM~\cite{lian2023unconstrained-udm} provides zero-order results. After the initial smoothed segmentation, we perform a 2D alignment search and further smoothed segmentation at the segment level. This yields 1st-order word segmentation and 1st-order word transcription. Additionally, we can use the 1st-order 2D-Alignment to guide the text refresher, which also provides us with 1st-order word transcription. We select the better of the two as the final 1st-order transcription, which is used as our final predictions. By repeating this process, we obtain 2nd-order word transcriptions, 3rd-order word transcriptions, and so on. For word segmentation evaluation, we utilize WhisperX~\cite{bain2023whisperx}, which provides timing information for each word. The results are detailed in Table \ref{word-trans-eval} for word transcription evaluation and Table \ref{word-seg-eval} for word segmentation evaluation. We also include disfluent speech segmentation results in the appendix~\ref{sec:appendix}.

\begin{table}[h]
    \centering
    \setlength{\tabcolsep}{10pt}
    \renewcommand{\arraystretch}{1.1} % Adjust the row spacing
    \resizebox{7cm}{!}{
        \small
        \begin{tabular}{l c c c c} 
            \toprule
            &\multicolumn{4}{c}{MS($\%$, $\uparrow$)} \\
            \hline
            URFA Config & Zero-order  & 1st-order  & 2nd-order & 3rd-order \\
            \hline
            Whisper-X & 42.1 & - &- &-\\
            \textbf{Ours} & \textbf{77.4} & \textbf{79.4} &\textbf{81.2} &\textbf{81.4}\\
            \bottomrule
        \end{tabular}
    }
    \caption{Word Segmentation Evaluation}
    \label{word-seg-eval}
\end{table}
% \vspace{-8mm}
\subsection{disfluency Detection}
We select UFA-VCTK and UFA-VCTK++ as the default phoneme transcriber, as they exhibit the best phonetic transcription performance, as demonstrated in Table \ref{phn-transcription-eval}. In this study, we also aim to investigate whether the proposed recursive inference algorithm can enhance disfluency detection. It's important to note that the representations used for disfluency detection are always based on the 2D-Alignment, but with different orders of computations, including 1st-order, 2nd-order, and 3rd-order. The results are presented in Table \ref{phn-disfluency-eval} and Table \ref{word-disfluency-eval}. MS refers to the "Matching Score," as explained in Section \ref{wer}.

\subsection{Results and Discussion}
\subsubsection{Transcription Analysis}
In the phonetic results presented in Table \ref{phn-transcription-eval}, UFA with VCTK/VCTK++ consistently outperforms the other baseline settings. Therefore, we only introduce monotonicity (CTC) to UFA+VCTK/VCTK++. Ultimately, the inclusion of CTC significantly enhances performance across all metrics. Regarding word transcription results, as shown in Table \ref{word-trans-eval}, we observe two aspects. First, when examining the default setting~\cite{lian2023unconstrained-udm}, which corresponds to the zero-order setting, we can see that CTC improves zero-order transcription results. Second, when we further explore recursive inference experiments, the results for the $(n+1)$th order are consistently better than those for the $n$th order. It's worth noting that CTC, which introduces monotonicity, further boosts performance. We have not yet explored scaling results, and we are unsure if this could yield a better scaling curve as shown in Fig.\ref{i-asr}. We leave this as a topic for future work. We refrained from investigating additional iterations, as performance tends to approach saturation. This observation aligns with the findings from Fig. \ref{URFA}, where, after the 1st-order URFA iteration, the detection of disfluent word boundaries surpasses that achieved in the zero-order iteration. This conclusion also holds true for disfluent word segmentation results, as reported in Table \ref{word-seg-eval}. Notably, our methods outperform those of\citet{bain2023whisperx} by a significant margin. Furthermore, we provide more examples in Appendix \ref{word-seg-fig-app} to illustrate its effectiveness.

\subsubsection{disfluency Analysis}
\begin{table}[h]
    \centering
    \setlength{\tabcolsep}{3pt}
    \renewcommand{\arraystretch}{1.5} 
     \resizebox{7.5cm}{!}{
     \small
    \begin{tabular}{l c c c c} 
     \toprule
    URFA Settings& F1 ($\%$, $\uparrow$) & MS ($\%$, $\uparrow$) & Human F1 ($\%$, $\uparrow$)& Human MS ($\%$, $\uparrow$)\\
     \hline
    UFA-VCTK&62.4&55.2&90.4&85.6\\
    UFA-VCTK$^{++}$&64.5&60.2&90.6&86.0\\
    \hspace{1em} +CTC&65.0&60.4&90.5&86.2\\
    \hspace{2em} +1st-order &65.6&61.0&90.6&86.0 \\
    \hspace{2em} +2nd-order &67.0&62.7&90.6&86.0 \\
    \hspace{2em} +3rd-order &\textbf{67.2}&\textbf{62.8}&\textbf{90.7}&\textbf{86.2} \\
    \bottomrule
    \end{tabular}}
    \caption{Phonetic disfluency Detection Evaluation}
    \label{phn-disfluency-eval}
\end{table}
We examine both phonetic-level and word-level dysfluencies in Table~\ref{phn-disfluency-eval} and Table~\ref{word-disfluency-eval}, respectively. It is evident that the introduction of CTC monotonicity consistently enhances performance at both levels. Additionally, when we consider recursive modeling, we can observe progressively improved performance as we increase the number of orders. However, it also reaches a point of saturation when we include further recursions.

\begin{table}[h]
    \centering
    \setlength{\tabcolsep}{10pt}
    \renewcommand{\arraystretch}{1.1} 
     \resizebox{8cm}{!}{
     \small
    \begin{tabular}{l c c} 
     \toprule
    Methods& F1 ($\%$, $\uparrow$) & Human F1 ($\%$, $\uparrow$)\\
     \hline
    Whisper-Large&64.0&86.4\\
    \hspace{1em} +Text Refresher(VCTK)&66.8&88.0\\
    \hspace{1em} +Text Refresher(VCTK$^{++}$)&68.4&89.1\\
    \hspace{2em} +CTC&68.8&89.2\\
    \hspace{3em} +1st-order &70.1&89.1\\
    \hspace{3em} +2nd-order &73.0&89.3\\
    \hspace{3em} +3rd-order &\textbf{73.1}&\textbf{89.3}\\
    \bottomrule
    \end{tabular}}
    \caption{Word disfluency Detection Evaluation}
    \label{word-disfluency-eval}
\end{table}

%% file: conclusion.tex
\section{Limitations}
We propose a hierarchical unconstrained dysfluency modeling (H-UDM), which is an extension of UDM~\cite{lian2023unconstrained-udm}. H-UDM introduces CTC monotonicity, and the incorporation of recursive modeling significantly enhances both transcription and disfluency detection results by a substantial margin. However, there are still several limitations to consider. First, the results on disordered speech are not as satisfactory. This suggests that the inference-only algorithm, the template matching algorithm, may not be sufficient for advanced disfluency modeling. It remains essential to develop end-to-end methods to address this challenge, which, however, presents its own set of difficulties. Second, the current closed-set disfluency classification only includes five types of disfluencies: replacement, insertion, repetition, block, and deletion. However, in an open-domain disfluency modeling system, there are many other complex disfluency patterns, such as syllable swapping and false starts. Designing specific templates for each type of disfluency is impractical. Third, phoneme units may not be the optimal choice for modeling disfluency. For instance, there may not be significant acoustic differences between "AH" and "AA," yet H-UDM treats them as two distinct phonemes. Although this is partially alleviated by smoothed segmentation, the improvement is limited. Therefore, it is worth exploring alternative speech units, such as articulatory units~\cite{lian22bcsnmf, lian2023factor, wu23k_interspeech, inversion}, to enhance alignment modeling.
\section{Acknowledgement}
Thanks for support from UC Noyce Initiative, Society of Hellman Fellows, NIH/NIDCD and the Schwab Innovation fund.

%% file: appen.tex
\section{Appendix}
\label{sec:appendix}

% \begin{figure}[h]
%     \centering
%     \includegraphics[height=7.5cm]{UFA.pdf}
%     \caption{UFA Module}
%     \label{ufa-fig}
% \end{figure}

\paragraph*{Human Data Annotation} \label{huamn-anno-app}
For all disordered speech (aphaisa and dylexia), our co-workers work together to manually label the dysfluencies: types of dysfluency and its time stamp at both word and phoneme level. As the dysfluency patterns are straightforward to observe, each utterance is labelled by only one person.

\section*{Word Segmentation Examples} \label{word-seg-fig-app}

GT denotes ground truth. Some samples might have multiple ground truths denoted as GT1, GT2, etc. 
\begin{figure*}[h]
    \centering
    \includegraphics[width=16cm, height=6cm]{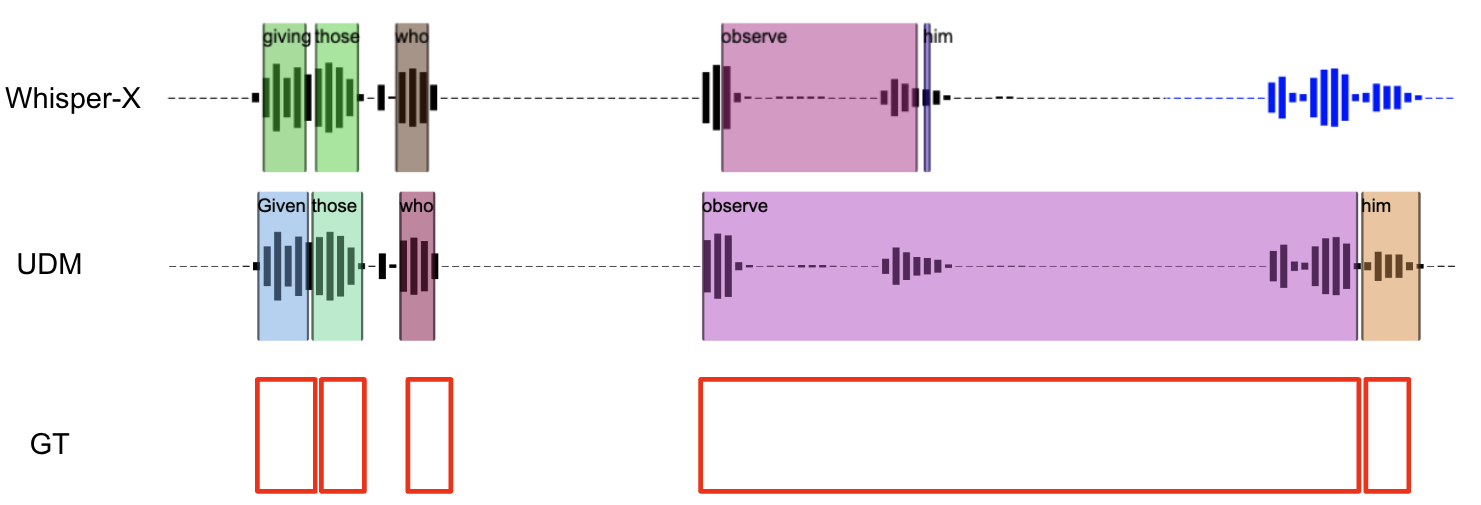}
    \caption[]{Segmentation-(Dyslexia Sample: Giving those who observe him)}
\end{figure*}

\begin{figure*}[h]
    \centering
    \includegraphics[width=16cm, height=6cm]{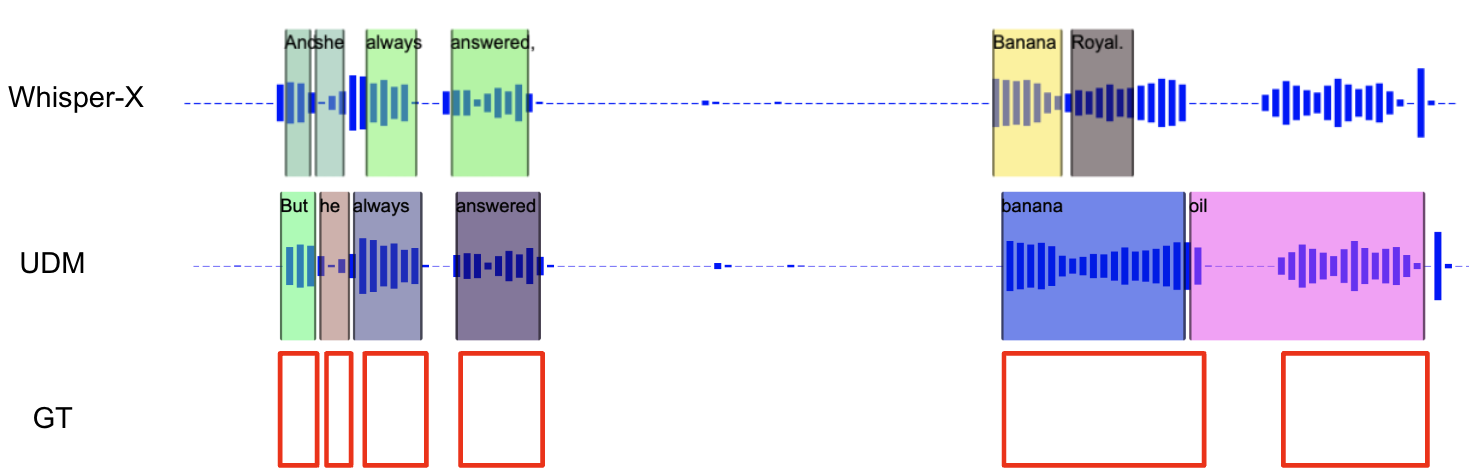}
    \caption[]{Segmentation-(Dyslexia Sample: But he always answered banana oil.)}
\end{figure*}

\begin{figure*}[h]
    \centering
    \includegraphics[width=16cm, height=6cm]{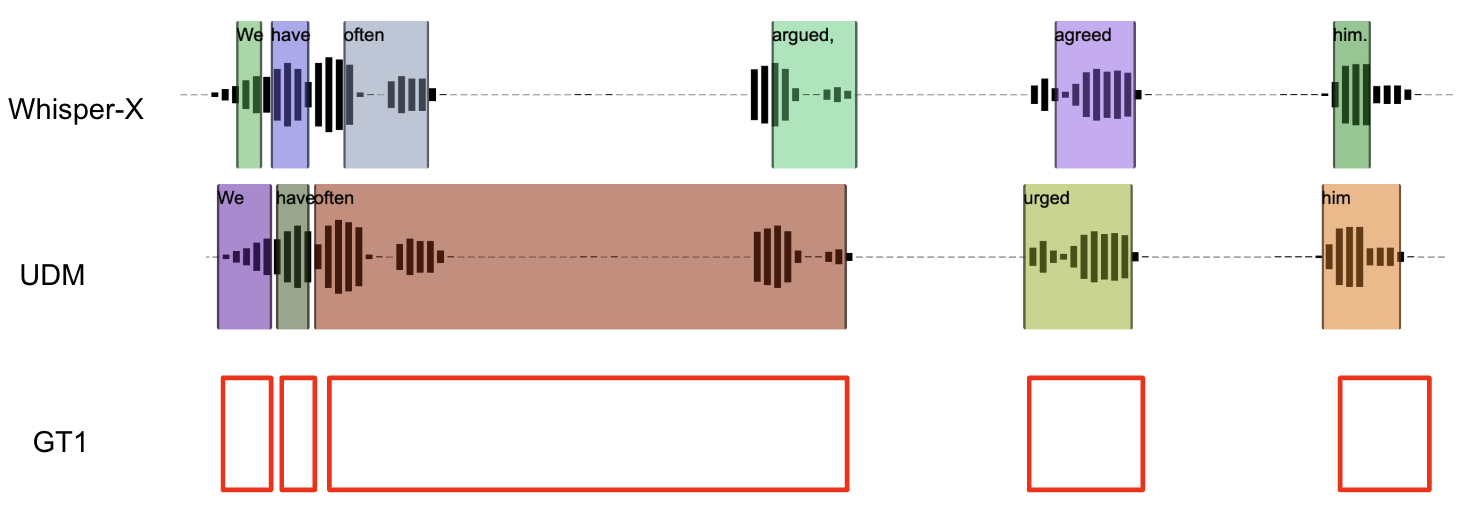}
    \caption[]{Segmentation-(Dyslexia Sample: We have often urged him)}
\end{figure*}

\begin{figure*}[h]
    \centering
    \includegraphics[width=16cm, height=6cm]{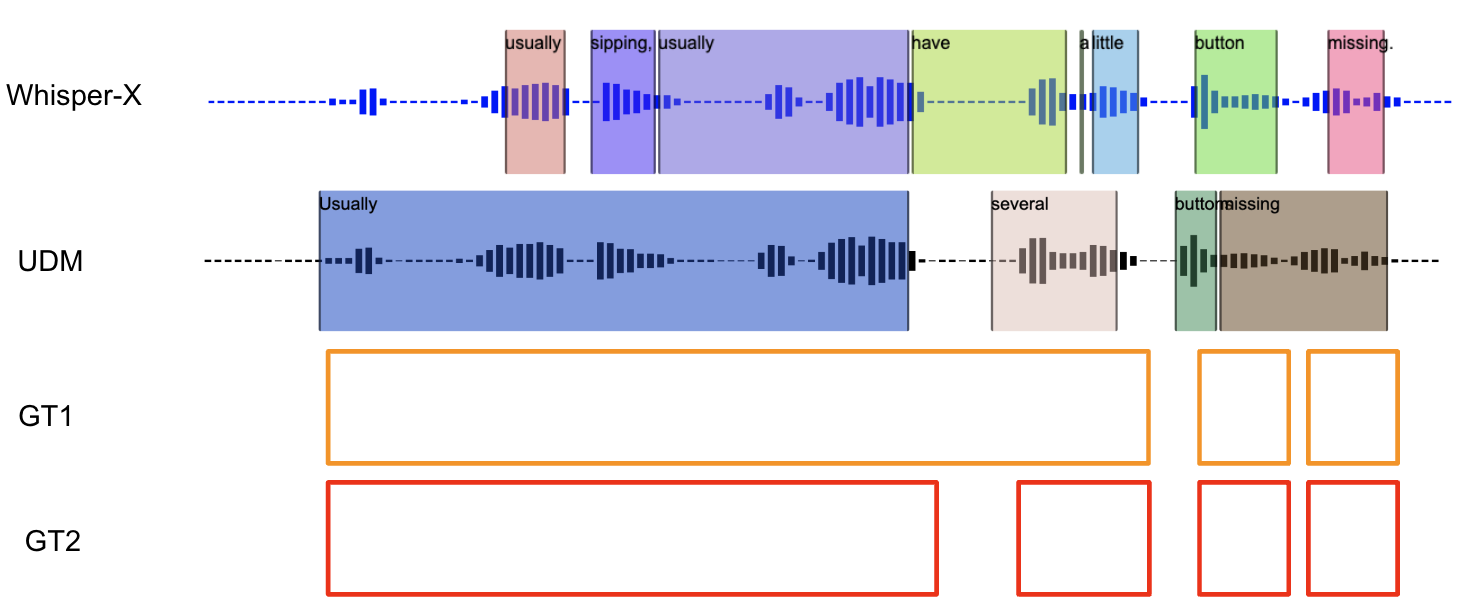}
    \caption[]{Segmentation-(Aphasia Sample: Usually several buttons missing.)}
\end{figure*}
\begin{figure*}[h]
    \centering
    \includegraphics[width=16cm, height=6cm]{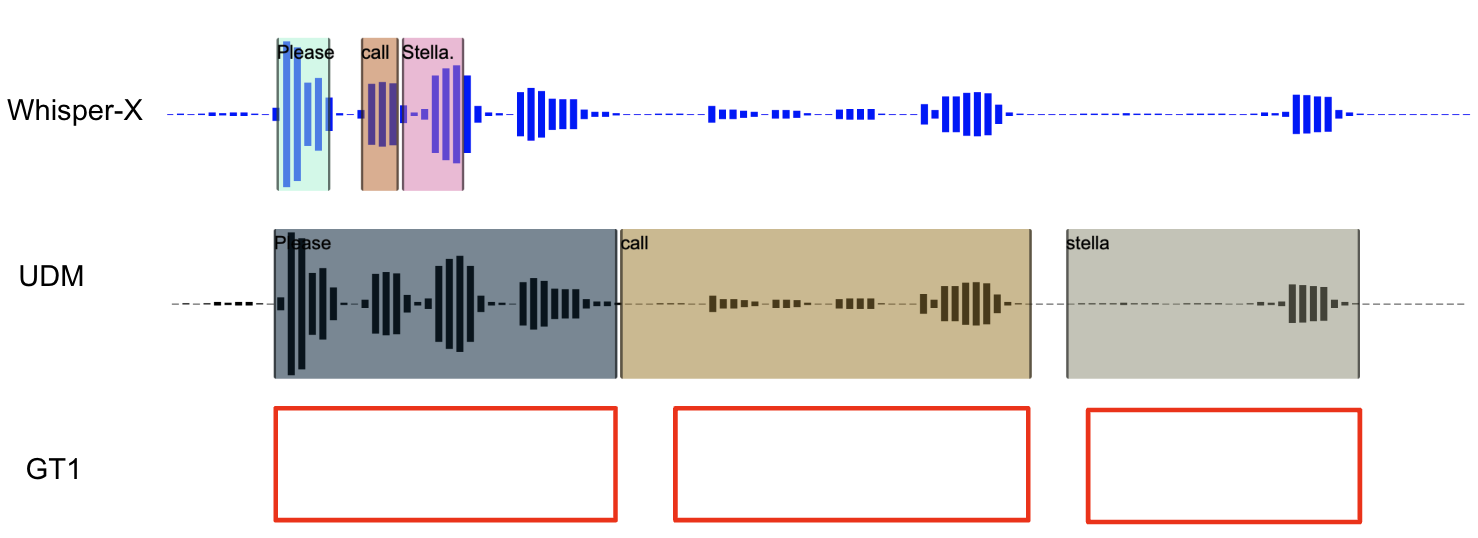}
    \caption[]{Segmentation-(My stutter sample: Please call stella.)}
\end{figure*}